\documentclass[letterpaper]{article} 
\usepackage{aaai2026}
\usepackage{booktabs}
\usepackage[breakable,most]{tcolorbox}
\newtcolorbox[auto counter]{conversation}[2][]
  {
   colback=gray!5.5!white,
   colframe=black!65!black, 
   fonttitle=\bfseries,
   fontupper=\sffamily\fontsize{7.5pt}{10.5pt}\selectfont,
   colbacktitle=gray!5.5!white, enhanced,
   coltitle=black,
   attach boxed title to top left={yshift=-2.5mm, xshift=4mm},
   title=#2, boxrule=0.3pt, #1,
   rounded corners, arc=2mm,
   boxed title style={boxrule=0.3pt, rounded corners, arc=2mm},
   label type=table
   }

\usepackage{times}  
\usepackage{helvet}  
\usepackage{courier}  
\usepackage[hyphens]{url}  
\usepackage{graphicx} 
\usepackage{caption}
\usepackage{subcaption}
\usepackage{natbib}  
\usepackage{multirow}
\usepackage{caption} 
\usepackage{algorithm}
\usepackage{algpseudocode}
\usepackage{longtable}
\usepackage{amsthm} 
\theoremstyle{plain}
\newtheorem{theorem}{Theorem}[section]
\newtheorem{proposition}[theorem]{Proposition}

\theoremstyle{definition}

\theoremstyle{remark}

\usepackage{newfloat}
\usepackage{listings}
\usepackage{amsmath}
\usepackage{amssymb}
\DeclareCaptionStyle{ruled}{labelfont=normalfont,labelsep=colon,strut=off} 
\floatstyle{ruled}
\newfloat{listing}{tb}{lst}{}
\floatname{listing}{Listing}
\title{Preference Robustness for DPO with Applications to Public Health}
\author{
    Cheol Woo Kim\equalcontrib,
    Shresth Verma\equalcontrib,
    Mauricio Tec\equalcontrib,
    Milind Tambe
}
\affiliations{
    Harvard University\\

}

\usepackage{bibentry}

\begin{document}

\maketitle

\begin{abstract}
We study an LLM fine-tuning task for designing reward functions for sequential resource allocation problems in public health, guided by human preferences expressed in natural language. This setting presents a challenging testbed for alignment due to complex and ambiguous objectives and limited data availability. We propose \textsc{DPO-PRO}, a robust fine-tuning algorithm based on Direct Preference Optimization (DPO), which accounts for uncertainty in the preference distribution using a lightweight Distributionally Robust Optimization (DRO) formulation. Unlike prior DRO-based variants, \textsc{DPO-PRO} focuses solely on uncertainty in preferences, avoiding unnecessary conservatism and incurring negligible computational overhead. We evaluate \textsc{DPO-PRO} on a real-world maternal mobile health program operated by the non-profit organization ARMMAN, as well as on standard alignment benchmarks. Experimental results demonstrate that our method consistently improves robustness to noisy preference signals compared to existing DPO variants. Moreover, \textsc{DPO-PRO} achieves comparable performance to prior self-reflection-based baseline for reward function design, while requiring significantly lower inference-time cost.
\end{abstract}


\section{Introduction}
\label{sec:intro}

Limited resource allocation is a core challenge in public health, where decision-makers must prioritize individuals for care or intervention under tight financial and operational constraints. In maternal healthcare, for instance, preventative care awareness programs have been shown to significantly reduce maternal mortality \cite{Helpmum, Jahan2018MomConnect, ARMMAN}, yet are often implemented by non-profit organizations that must serve large populations with limited resources. To support such efforts, it is crucial to develop allocation policies that make effective sequential decisions based on individual-level attributes such as age, income, or health status.

Reinforcement learning (RL) offers a powerful framework for learning allocation policies in complex, dynamic environments. A widely used model in this context is the Restless Multi-Armed Bandit (RMAB)~\cite{math11071639}, a class of Markov Decision Processes (MDPs) well-suited for sequential decision-making under resource constraints. However, in RL, it is well-recognized that designing an appropriate reward function is critical \cite{singh09,Booth_Knox_Shah_Niekum_Stone_Allievi_2023}. In public health applications, decision-makers' priorities (e.g., which populations to prioritize) must be encoded into the reward function, so that resulting policy outcomes align with long-term public health goals. Crafting a reward function that accurately captures these priorities is often labor-intensive, demanding manual tuning, repeated simulation, and domain expertise. As public health goals shift over time, this design process must be repeated, limiting the scalability and adaptability of RL-based solutions in dynamic, real-world environments.

Recently, several works explored using large language models (LLMs) to automate reward function design for sequential decision-making problems \citep{mirchan21,ijcai2019p331, carta2022eager,ma2024eureka, behari2024a, verma2025balancingactprioritizationstrategies}. In these works, an LLM interprets a user's high-level objective (e.g., desired policy outcomes) in natural language and outputs a reward function that guides RL policy training. This enables efficient workflow for developing human-aligned RL policies. Recent studies demonstrated that LLMs can effectively generate reward functions for RMAB problems in public health domains \citep{behari2024a, verma2025balancingactprioritizationstrategies}.

Existing approaches often rely on inference-time optimization via self-reflection \cite{shinn2023reflexion}, where the LLM iteratively refines its outputs based on simulated feedback. Although this strategy provides flexibility, it incurs substantial inference-time costs due to repeated rounds of simulation and reasoning. In many non-profit settings, data sensitivity constraints further limit the use of large proprietary models, pushing organizations toward open-source alternatives, which are often smaller and less capable. Because self-reflection is highly dependent on model strength, its effectiveness can be diminished in these environments. As a result, fine-tuning open-source models offers a more practical path: it directly improves their capabilities while avoiding the high inference costs and privacy issues in large-scale public health deployments \cite{wu2024inference}. This makes fine-tuning particularly suitable for resource-constrained domains such as public health \cite{zhao2025foundation}.

Thus, we aim to move beyond inference-time methods and instead fine-tune an LLM to more effectively translate natural language preferences into high-performing reward functions for RMABs. A promising approach is Direct Preference Optimization (DPO) \citep{rafailov2023DPO}, which enables LLM fine-tuning from pairwise preference annotations. Recent work has also leveraged LLM-as-a-judge to scale annotation as a proxy for human feedback~\cite{zhang2025aaai, tan-etal-2024-large, ultra2024, zhu2024starlingb}. However, applying these techniques to the public health domain introduces significant challenges, particularly due to the presence of noise in the preference signal.


\textit{First}, the objectives in the public health domain are often ambiguous or subjective (e.g., ``prioritize older individuals'').
\textit{Second}, reward design in RMABs demands complex reasoning, as the quality of a proposed reward function is not evaluated in isolation, but rather by the policy it induces, increasing the complexity of assessing preference.  
\textit{Third}, real-world public health datasets are typically smaller, increasing the risk of misalignment due to overfitting and reward hacking, which is a common problem with reinforcement learning for human feedback methods such as DPO \citep{pmlr-v238-gheshlaghi-azar24a, amini-etal-2024-direct, yang2024orthogonalfinetuningdirectpreference, xiao2025comprehensivesurveydirectpreference}. However, public health is a high-stakes domain, where poorly aligned policies can lead to serious real-world consequences.

{To tackle these challenges, we introduce \textsc{DPO-PRO}, a lightweight and distributionally robust enhancement of DPO. \textsc{DPO-PRO} incorporates uncertainty in the preference distribution using an efficient yet effective Distributionally Robust Optimization (DRO) formulation \citep{Rahimian_2022}, ensuring that the fine-tuned policy remains reliable under imperfect preference annotation.} Unlike prior DRO-based DPO methods \citep{wu2025drdpo, xu2025distributionallyrobustdirectpreference, mandal2025distributionallyrobustreinforcementlearning}, our approach avoids excessive conservatism and only adds negligible computational overhead. We further show that the resulting robust loss can be interpreted as a regularized DPO loss that penalizes model overconfidence and weak preference signals.

We evaluate \textsc{DPO-PRO} on the UltraFeedback benchmark \cite{ultra2024} and on a simulated maternal health task based on ARMMAN’s real-world mobile health program \cite{ARMMAN}. In this setting, health workers can place only a limited number of live service calls each week to boost beneficiary engagement. Prior work models this problem using RMABs, where each beneficiary’s engagement evolves over time and the objective is to allocate a limited number of calls to maximize long-term engagement. Our experiments show that \textsc{DPO-PRO} is more robust to noisy preference annotation than vanilla DPO and prior DRO-based approaches, both on UltraFeedback and across a range of public health objectives. We further show that \textsc{DPO-PRO} achieves performance comparable to the Decision Language Model (DLM), a self-reflection–based method used in the ARMMAN setting \cite{behari2024a}, while requiring significantly lower inference-time cost (See Figure~\ref{fig:inf} in Section \ref{sec:benchmarks-health}).

\section{Related Works}
\label{sec:related}

A growing body of work has explored robustness in RLHF and DPO, particularly by leveraging DRO. \citet{wu2025drdpo} introduce a DRO-based DPO formulation using a KL-divergence-based ambiguity set. \citet{xu2025distributionallyrobustdirectpreference} extend this approach to both KL and Wasserstein distances, while \citet{mandal2025distributionallyrobustreinforcementlearning} explore robustness under total variation distance, applying it to both DPO and standard two-stage RLHF.

In contrast to these DRO-based approaches, other works assume explicit corruption models. For example, \citet{chowdhury2024} assume that preference labels are flipped according to a known noise rate, and \citet{bukharin2024robust} model preferences using a Bradley–Terry (BT) framework \citep{BT} with added noise on the reward differences.

\citet{hong2024adaptive} use a DRO framework to develop adaptive loss functions for reward model learning, but focus is not on distributional robustness. \citet{zhan2024provable} introduce robustness by constructing confidence sets over learned reward functions, followed by pessimistic policy optimization.

Broadly, existing robustness techniques fall into two categories: (1) methods relying on strong assumptions about noise models, which may be unrealistic in practice, and (2) DRO-based methods that avoid explicit corruption modeling but hedge against worst-case distributions. As discussed in Section \ref{subsec:offdpo}, standard DRO approaches tend to be overly conservative and often require solving challenging min–max optimization problems. For tractability, prior methods rely on heuristic approximations \citep{wu2025drdpo, xu2025distributionallyrobustdirectpreference}, potentially undermining theoretical guarantees or practical robustness. In contrast, we provide efficient approach DPO-PRO, with theoretical guarantees, and simultaneously illustrate its applicability in real world public health settings.

\section{Preliminaries}
\label{sec:prelim}

\subsection{Data Distribution}
\label{subsec:data}

In DPO, each data point is a tuple $(x,y_1,y_2,c)$. A prompt $x \in \mathcal{X}$ is drawn from a distribution $\mu$. Given $x$, responses $y_1,y_2 \in \mathcal{Y}$ are sampled independently from a policy $\pi(\cdot|x)$. A label $c \in \{1,-1\}$ is a Bernoulli random variable indicating whether response $y_1$ is preferred over $y_2$, with preference probability given by $p^*(y_1 \succ y_2 |x)$. The distribution \( p^* \) represents the ground-truth preference distribution that the user intends the language model to align with.

In summary, the data-generating process involves three components: (1) the prompt distribution, (2) the response distribution, and (3) the preference distribution. This yields the following joint distribution:
$
P(x,y_1,y_2,c)
  \;=\;
  \mu(x)\,
  \pi(y_1|x)\,
  \pi(y_2|x)\,
  \bigl[
      \mathbb{I}_{\{c=1\}}\,p^{*}(y_1 \succ y_2 \mid x)
      \;+\;
      \mathbb{I}_{\{c=-1\}}\,p^{*}(y_2 \succ y_1 \mid x)
  \bigr].
$
We often omit the dependence of $p^*$ (and other preference distributions) on $(x,y_1,y_2)$ and simply write $p^*$ instead of $p^*(y_1 \succ y_2 |x)$ when the meaning is clear from context.

\subsection{(Distributionally Robust) DPO}
\label{subsec:offdpo}

In standard RLHF, the first step is to learn a reward function $R_{\phi}$ from a fixed dataset $\mathcal{D}$. This is typically done by minimizing the negative log-likelihood:
\begin{equation*}
-\mathbb{E}_{(x, y_1, y_2, c) \sim \mathcal{D}} \left[
  \log \sigma\left( c \cdot \left( R_\phi(x, y_1) - R_\phi(x, y_2) \right) \right)
\right],
\end{equation*}
where $\sigma$ is the sigmoid function.

Once the reward model is trained, a policy $\pi_{\theta}$ is optimized to maximize the expected reward while remaining close to a reference policy $\pi_{ref}$, typically via Proximal Policy Optimization (PPO) \cite{schulman2017ppo}:
\begin{equation*}
\max_{\pi_\theta} \mathbb{E}_{(x,y)\sim\pi_\theta} \left[ R_\phi(x, y) \right] 
- \beta\, \mathrm{KL}(\pi_\theta \,\|\, \pi_{\mathrm{ref}}),
\end{equation*}
where $\beta$ is a regularization parameter controlling the strength of the KL penalty.

DPO simplifies this pipeline by combining the two stages into a single objective that directly optimizes the policy from pairwise preference data. The DPO loss $\mathcal{L}_{\text{DPO}}(\pi_\theta)$ is given by:
\begin{equation*}  
-\mathbb{E}_{(x,y_1,y_2,c)\sim\mathcal{D}}\!\bigl[
  \log\sigma\!\bigl(
      c\,\beta\Bigl(
        \log\tfrac{\pi_\theta(y_1|x)}{\pi_{\mathrm{ref}}(y_1|x)}
        - \log\tfrac{\pi_\theta(y_2|x)}{\pi_{\mathrm{ref}}(y_2|x)}
      \Bigr)
  \bigr)
\bigr].
\end{equation*}

\noindent For brevity, we define the per-sample loss as:
$$
\ell_\theta(x,y_1,y_2,c)=
-\log\sigma\!\Bigl(
       c\,\beta\!\Bigl(
         \log\tfrac{\pi_\theta(y_1|x)}{\pi_{\mathrm{ref}}(y_1|x)}
         - \log\tfrac{\pi_\theta(y_2|x)}{\pi_{\mathrm{ref}}(y_2|x)}
       \Bigr)\Bigr).
$$
Using this expression, $\mathcal{L}_{\text{DPO}}(\pi_\theta)$ is also equivalent to 
$$
\mathbb{E}_{(x,y_1,y_2)\sim\mathcal{D}}[p^*\ell_\theta(x,y_1,y_2,1) + (1-p^*)\ell_\theta(x,y_1,y_2,-1)].
$$
When the context is clear that the triplet $(x,y_1,y_2)$ and the parameter $\theta$ is fixed, we also use $\ell_1$ and $\ell_{-1}$ to denote $\ell_\theta(x,y_1,y_2,1)$ and $\ell_\theta(x,y_1,y_2,-1)$, respectively. 

As discussed earlier, DPO is vulnerable to overfitting and noise in the data. To address this, recent work has proposed distributionally robust objectives of the form $\max_{P \in \mathcal{Q}(\mathcal{D})}\mathcal{L}_{{P}}(\pi_\theta),$
where $\mathcal{Q}(\mathcal{D})$ denotes an ambiguity set centered around the empirical distribution (data) $\mathcal{D}$ and the loss $\mathcal{L}_{{P}}$ is the DPO loss evaluated under the perturbed distribution $P$ rather than the original data distribution \citep{xu2025distributionallyrobustdirectpreference, wu2025drdpo}.

\paragraph{Conservatism and Computational Costs}
The above DRO formulation hedges against shifts in the entire joint distribution on $(x,y_1,y_2,c)$. This allows the adversary to assign weights to highly unlikely prompts or responses. The outer minimization must then optimize against losses in these practically irrelevant regions, which might make the update overly conservative and slow improvement on the data the model actually sees.  From a computational standpoint, solving the resulting min–max problem can be intractable in practice, and existing methods often rely on approximations that deviate from the original DRO formulation.

\subsection{Reward Function Design Tasks for RMABs}
\label{subsec:rewarddesign}

In our setting, the prompt $x$ represents a desired policy outcome specified by a human prompter, for example, \texttt{Focus on the young mothers by age and also focus on those with low income}. The responses $y_1$ and $y_2$ correspond to candidate reward functions for RMABs. The role of the LLM-annotator is to judge which of the two reward functions, $y_1$ or $y_2$, better aligns with the objective expressed in the prompt.

A key challenge in this task is that the quality of a reward function cannot be assessed in isolation. Rather, it must be evaluated based on the policy it induces when optimized, i.e., how effectively the resulting policy fulfills the user’s (often ambiguous) intent. This requires the annotator to infer the downstream effects of each reward function. Given the complexity and ambiguity involved, the resulting preference signals are inevitably noisy and should not be treated as fully reliable.

\section{DPO with Preference Robustness}
\label{sec:main}

In this section, we introduce \textsc{DPO-PRO}, a distributionally robust version of DPO that specifically targets distributional shifts or other forms of noise in the preference distribution. The complete proofs for all theorems are provided in the extended version of this paper \cite{kim2025dpopro}, which includes the full technical appendix.



\subsection{Uncertainties in the Preference Distribution}
\label{subsec:dro}

In each data point $(x,y_1,y_2,c)$, the triplet $(x,y_1,y_2)$ represents observed content, which typically comes from a well-understood and controllable data collection process. In general, $x$ is drawn from a pool of candidate inputs, and the response pair $(y_1,y_2)$ is generated by the reference policy $\pi_{ref}$ (usually a supervised-fine tuned model) for every sampled prompt $x$. As more data is collected, the uncertainty in this part of the data is expected to diminish. Noise in the prompts and responses becomes even less of an issue under iterative DPO methods, which are increasingly adopted in practice \cite{xiong2024iterative, cen2025valueincentivized, xie2025exploratory}.

In contrast, we argue that the preference distribution $c \sim p^* $ is inherently noisy in practice, and collecting more data is unlikely to resolve the underlying uncertainty in human preference. Even when humans are directly annotating, various sources of noise remain, including human subjectivity, irrational or inconsistent behavior, and temporal variability (e.g., the same annotator providing different judgments at different times). This noise is persistent and cannot be easily addressed through better data curation or increased data volume. In other words, the ground-truth preference distribution $p^*$ is an idealized and inaccessible object in practice. As mentioned in the introduction, this challenge is particularly pronounced in public health domains, where objectives are often ambiguous or subjective (e.g., ``prioritize older individuals"), and reward design demands complex reasoning about long-term policy consequences that are difficult to assess accurately. Moreover, real-world public health datasets are typically smaller, further exacerbating preference uncertainty and increasing the risk of misalignment. Therefore, we argue that uncertainty in the preference distribution should be the primary motivation for incorporating robustness into DPO.

We assume that for each prompt and response pair $(x,y_1,y_2)$, we have access to a (potentially noisy) preference distribution $q(y_1 \succ y_2 |x)$. We apply DRO adjustment, assuming worst-case deviation from $q$ within a pre-specified chi-squared divergence ball.

Formally, we define the DRO loss $\mathcal{L}_{\text{DRO}}(\pi_\theta)$ as:
\begin{equation}
 \mathbb{E}_{(x,y_1,y_2)\sim\mathcal{D}}
    \max_{p \in Q(x,y_1,y_2,\rho)}
    \mathbb{E}_{c \sim p}\Bigl[
      \ell_\theta(x,y_1,y_2,c)
    \Bigr]
\label{eq:dro_loss}
\end{equation}
where $Q(x,y_1,y_2 \rho)=\left\{ p : \chi^2\big(p \,\|\, q(y_1 \succ y_2 \mid x)\big) \leq \rho \right\}$. Our formulation avoids unnecessary robustification over the joint distribution on $(x,y_1,y_2)$, and assumes that this part of the data is reliable. Furthermore, unlike standard DRO methods that can add significant computational burden \cite{Rahimian_2022}, our formulation introduces negligible additional cost as shown next.

\subsection{Computing Worst-case Distribution}
\label{subsec:worst}

Consider a single data point $(x,y_1,y_2)$. We assume access to an estimate of the probability that one response is preferred over the other: $q(y_1 \succ y_2 |x) \in (0,1)$.  

Given this probability $q$, we define the following per-sample optimization problem: 
\begin{equation}
\begin{aligned}
\max_{p\in[0,1]} \quad
& 
  p\,\ell_{\theta}(x,y_{1},y_{2},1)
  +(1-p)\,\ell_{\theta}(x,y_{1},y_{2},-1) \\[3pt]
\text{s.t.}\quad
& \frac{(p-q)^2}{q(1-q)}\;\le\;\rho
\end{aligned}
\label{eq:worst}
\end{equation}
This problem finds the worst-case distribution within the chi-squared ambiguity set centered at $q$, $\left\{ p : \chi^2\big(p \,\|\, q(y_1 \succ y_2 \mid x)\big) \equiv \frac{(p-q)^2}{q(1-q)}\leq \rho \right\}$. Let $\hat{p}(y_1 \succ y_2 |x)$ denote the optimal solution to this problem. Using this, the robust loss $\mathcal{L}_{\text{DRO}}(\pi_\theta)$ can be equivalently expressed as 
$$ \mathbb{E}_{(x,y_1,y_2)\sim\mathcal{D}}\Bigl[\hat{p}
      \ell_\theta(x,y_1,y_2,1) + (1-\hat{p})
      \ell_\theta(x,y_1,y_2,-1)
    \Bigr].$$

Eq \eqref{eq:worst} is a one-dimensional optimization problem with a linear objective, which admits a simple closed-form solution:
\begin{equation*}
\hat{p}(y_1 \succ y_2 |x) =
\begin{cases}
\min\left\{1,q + \sqrt{\rho\, q(1 - q)}\right\}, \ \text{if } \ell_{1} \ge \ell_{-1}, \\[6pt]
\max\left\{0,q - \sqrt{\rho\, q(1 - q)}\right\}, \ \text{if } \ell_{1} < \ell_{-1}.
\end{cases}
\end{equation*}

Consider the case where \( \ell_1 \ge \ell_{-1} \), which implies that under the current policy \( \theta \), the model assigns a higher likelihood to \( y_2 \) being preferred over \( y_1 \). To increase the loss, the adversary seeks to shift the preference probability \( q(y_1 \succ y_2 \mid x) \) in the opposite direction. That is, it tries to increase $q$ to emphasize a preference for \( y_1 \), which contradicts the model’s belief. Consequently, the worst-case distribution becomes \( q + \sqrt{\rho\, q(1 - q)} \). Since probabilities must remain within the unit interval, we clip the value at 1, yielding the final expression \( \min\left\{1,\; q + \sqrt{\rho\, q(1 - q)}\right\} \).

\subsection{Efficient Optimization of the Loss}
\label{subsec:loss_compute}

The worst-case probability obtained from the optimization problem above, $\hat{p}$, replaces the preference probability in the per-sample DPO gradient as well:
\begin{equation}
\hat{p}\,\nabla_\theta\ell_{\theta}(x,y_1,y_2,1) \;+\; (1-\hat{p})\,\nabla_\theta\ell_{\theta}(x,y_1,y_2,-1).
\label{eq:grad_est}
\end{equation}

\begin{proposition}
\label{prop:dro}
    Eq \eqref{eq:grad_est} provides an unbiased gradient estimate of the DRO loss in Eq \eqref{eq:dro_loss}.
\end{proposition}
Note that we do not formally differentiate through the inner maximization in Eq \eqref{eq:dro_loss} with respect to $\theta$. However, Danskin's theorem \citep{Bertsekas99Np} justifies our approach. For a fixed $\theta$, we may solve the inner maximization and directly substitute the resulting worst-case $\hat{p}$ into the gradient expression. 

We emphasize the resulting DRO gradient is both exact and computationally efficient. Unlike prior work, our approach does not introduce any approximation or heuristic in minimizing the DRO objective. Moreover, our distributional robustness is applied only to the preference distribution $q$, rather than the full data-generating distribution on $(x,y_1,y_2,c)$. Hence, the resulting method is significantly less conservative than earlier DRO-based DPO approaches, and matching more accurately the type of distribution shift we expect in the true human preferences in our public health application as seen later.

\subsection{Analysis of the Loss Function}
\label{subsec:loss}

In this section, we analyze the robust loss \( \mathcal{L}_{\text{DRO}} \) and show that it is equivalent to the original DPO loss augmented with a regularization term. This interpretation provides insight into how the DRO formulation affects model learning.

\begin{proposition}
\label{prop:loss_regular}
The DRO loss $\mathcal{L}_{\text{DRO}}$ is equivalent to regularizing the original DPO loss $\mathcal{L}_{\text{DPO}}$ as:
\begin{equation*}
\mathcal{L}_{\text{DRO}} =
\begin{cases}
\mathcal{L}_{\text{DPO}} + \min\left\{1 - q,\; \sqrt{\rho\, q(1 - q)}\right\} (\ell_1 - \ell_{-1}), \\[4pt]
\hspace{3em} \text{if } \ell_1 \ge \ell_{-1}, \\[8pt]
\mathcal{L}_{\text{DPO}} + \min\left\{q,\; \sqrt{\rho\, q(1 - q)}\right\} (\ell_{-1} - \ell_1), \\[4pt]
\hspace{3em} \text{if } \ell_1 < \ell_{-1}.
\end{cases}
\end{equation*}
\end{proposition}

 We now analyze the behavior of the regularization term, assuming that $\ell_1 \geq \ell_{-1}$. The opposite case follows symmetric logic.

\paragraph{(1) Uncertainty-Weighted Coefficient.}  
The first factor \( \min\left\{1 - q,\; \sqrt{\rho\, q(1 - q)}\right\} \) penalizes cases where the preference signal is uncertain (i.e., when \( q \) is close to 0.5). This term is largest when \( q \) is near 0.5 and diminishes as \( q \to 0 \) or \( q \to 1 \). Its magnitude is modulated by the robustness radius \( \rho \), with larger \( \rho \) leading to a stronger penalty.


\paragraph{(2) Model Confidence.}  
The second factor \( (\ell_1 - \ell_{-1}) \) reflects the log-odds of the current model's preference for \( y_2 \) over \( y_1 \). A large value indicates strong model confidence. This term increases when the model becomes more certain about the preference, regardless of the ground-truth label.

\paragraph{(3) Combined Effect.}  
The regularization term penalizes the model for being overly confident in its preferences when the preference signal \( q \) is ambiguous. In contrast, when the signal is strong (i.e., \( q \) is close to 0 or 1), the penalty diminishes, allowing the model to express stronger preferences. In effect, the DRO loss encourages calibrated learning: the model is allowed to be confident only when the preference signal is also confident.

\subsection{Obtaining the Soft Score $q$ in Practice}
The DRO adjustment described above requires a soft score $q(y_1 \succ y_2 |x)$, rather than a binary annotation for each prompt and response pair $(x,y_1,y_2)$. Soft scores can be obtained from LLM judges in several ways: (1) producing multiple binary preference judgments through repeated queries and averaging the outcomes; (2) directly outputting a numerical preference score via appropriate prompting; or (3) extracting the log-probabilities assigned to tokens representing the preferences (e.g., \texttt{1}  for $y_1 \succ y_2$ and \texttt{-1}  for $y_2 \succ y_1$), and applying a softmax transformation to estimate a preference distribution \cite{lee2024}. 
When using a off-the-shelf reward model, the scalar outputs of the reward model can be converted into a pairwise preference probability using the BT model. We provide further discussion on annotation formats and the connection to robustness in \cite{kim2025dpopro}.

\section{Experiments}
\label{sec:exp}

We validate \textsc{DPO-PRO} in two different settings. 
First, we validate the methodology on a standardized benchmark for preference-based LLM finetuning. Next, we adapt this experiment for the ARMMAN  maternal health dataset \cite{ARMMAN} in the following section. The complete experimental details are provided in the extended version of this paper \cite{kim2025dpopro}.

\subsection{Benchmarks on General Alignment Datasets}\label{sec:benchmarks-u0}
 
We use the recent UltraFeedback alignment benchmark dataset \cite{ultra2024}, consisting of  60,000 top-quality human pairwise preference annotations that cover a wide variety of tasks and instruction types, rendering it a broad benchmark for examining alignment research. 

\paragraph{Noise in the training data}  Our primary objective is to evaluate how effectively preference-based LLM fine-tuning methods handle shifts in preference distributions between training and evaluation data. We use $q^*$ denote the true preference distribution used at test time. For our experiments, we define $q^*$ using the state-of-the-art preference reward model Eurus 7B\citep{yuan2025eurus}, which has been trained for the UltraFeedback dataset \citep{ultra2024}. We compute $q^*$ using the BT model. 

To simulate noise during training, we construct a noisy preference distribution $q_\alpha$ for training as the mixture
\begin{equation}\label{eq:noise-q}
q_\alpha := q^* (1 - \alpha) + (1 - q^*) \alpha.
\end{equation}
Intuitively, $q_\alpha$ ``flips" the preference distribution with probability $\alpha$. We refer to this type of noise interchangeably as \textit{label switching}. During training, we have access to $q_\alpha$ but not to $q^*$, with $\alpha$ being unknown. 

\paragraph{Training details} We base our experiments on the recent Phi 3-mini 3B model without instruction \cite{abdin2024phi3}. We initially train for two epochs of supervised fine tuning (SFT), followed by one epoch of DPO, with the baselines discussed below. All experiments are performed on a Nvidia A100 GPU 40 GB with batch size 2. 

\paragraph{Baselines}
We compare \textsc{DPO-PRO} against the standard DPO~\citep{rafailov2023DPO} and the prior distributionally robust variant DrDPO~\citep{wu2025drdpo} using their best hyperparameters suggested by the authors. We evaluate \textsc{DPO-PRO} under several values of the robustness parameter $\rho$ based on the $\chi^2$-divergence. Models are trained under two levels of label-flip noise, $\alpha=0.3$ (low) and $\alpha=0.6$ (high), as well as the noiseless setting $\alpha=0$.

\paragraph{Evaluation}
After the fine-tuning step, we generate a response $y_g \sim \pi_\theta(x)$ for every prompt $x$ in the evaluation dataset and compute its reward $R_\phi(x, y_g)$. We consider two evaluation metrics. The first is the \emph{win rate}, computed by comparing the generated response with the chosen response $y_c$ in the dataset using an LLM-as-a-judge approach with GPT-4o-mini~\cite{openai_gpt4omini_2024}. To mitigate position bias, we randomly flip the order of the two responses presented to the judge. The second metric, \emph{evaluation reward}, is computed as the average reward score defined as
$\frac{1}{N_\mathrm{eval}}\sum_{i=1}^{N_\mathrm{eval}} R_\phi(x, y_g).
$

\paragraph{Results}
Table~\ref{tab:bt-noise-flip-combined} summarizes the results. For the LLM-as-a-judge evaluation, the different \textsc{DPO-PRO} variants (each corresponding to different $\rho$ value) tend to underperform the baselines in the low-noise setting, but their relative performance improves as noise increases. Under high noise, all \textsc{DPO-PRO} variants outperform both vanilla DPO and DrDPO. The effect of~$\rho$ follows the expected pattern: smaller~$\rho$ values perform well when noise is low but degrade more quickly as noise increases, while larger~$\rho$ values exhibit more stability. Overall, the \textsc{DPO-PRO} family shows improved robustness as the preference noise becomes larger.

For metrics based on the reward model, \textsc{DPO-PRO} outperforms the baselines even in the noiseless setting, except for the case with $\rho = 0.1$. As expected, performance declines for all methods as noise increases. In the low-noise regime, however, the \textsc{DPO-PRO} variants perform slightly worse than the baseline, but they regain and ultimately surpass baseline performance as the noise becomes even larger. Note that this does not contradict the DRO objective, because the true optimization target for fine-tuning is reward plus KL regularization, while our evaluation here relies solely on reward-model scores. Moreover, these scores are only a proxy for quality, particularly for out-of-distribution responses, and are limited by the reward model’s own generalization capabilities.

\begin{table}[h]
\small
\centering

\begin{subtable}[t]{.99\columnwidth}
\centering
\caption{Win rate (\%) based on LLM-as-a-judge evaluation ($\uparrow$).}
\scalebox{0.85}{
\begin{tabular}{lcccc}
\toprule
Method & No noise  & Low noise  & High noise  \\
\midrule
DPO-PRO ($\rho=0.008$) & 43.24 & 35.17 & 28.75 \\
DPO-PRO ($\rho=0.03$)  & 41.71 & 33.33 & 28.98 \\
DPO-PRO ($\rho=0.1$)   & 40.33 & 33.54 & 27.97 \\
DPO                     & 43.77 & 36.19 & 26.97 \\
DrDPO                   & 44.85 & 38.84 & 25.51 \\
\bottomrule
\end{tabular}}
\end{subtable}


\begin{subtable}[t]{.99\columnwidth}
\centering
\caption{Evaluation reward ($\uparrow$).}
\scalebox{0.85}{
\begin{tabular}{lcccc}
\toprule
Method & No noise  & Low noise  & High noise \\
\midrule
DPO-PRO ($\rho=0.008$) & 923.70 & 713.00 & 446.65 \\
DPO-PRO ($\rho=0.03$)  & 912.65 & 695.00 & 499.59 \\
DPO-PRO ($\rho=0.1$)   & 781.28 & 656.32 & 499.36 \\
DPO                     & 877.01 & 729.12 & 400.44 \\
DrDPO                   & 897.67 & 753.77 & 419.05 \\
\bottomrule
\end{tabular}}
\end{subtable}

\caption{Results on UltraFeedback under label-flip noise.}
\label{tab:bt-noise-flip-combined}

\end{table}

\subsection{Benchmarks on Maternal Health Data}
\label{sec:benchmarks-health}

We apply our proposed method in the context of mobile health program run by \citet{ARMMAN}. We use real-world anonymized beneficiary listenership data from a quality-improvement study conducted by ARMMAN\footnote{The authors have been access to this restricted dataset under a data usage agreement.}. The data consists of registration information of beneficiaries (mothers) containing their socio-demographic features of age, income education as well as their preferences on call slot times and language. Additionally,
we have information on beneficiaries' interaction with automated voice calls and live service calls made by health-workers. We model the problem of allocating live service calls as a RMAB problem. For all our experiments, we consider a population with $2100$ beneficiaries (i.e., 2100 arms) and the number of live service calls that can be made every week is limited to 210. 

We use the Phi 3-Mini 3B model, and the noise-injection and training procedures follow the same setup as in Section~\ref{sec:benchmarks-u0}. Here, we experiment only with the noiseless and low-noise settings to better reflect realistic conditions, rather than injecting artificially large noise.


\paragraph{Preference Dataset Construction} 
We generate a preference dataset of health worker prioritization commands, preferred, and rejected reward functions. Since it is costly to generate this dataset through human annotation, we use an LLM judge with ChatGPT 4o-mini \citep{openai_gpt4omini_2024} as follows: i) query LLM-judge to obtain 20 candidate reward functions that align with the prioritization command, ii) for each candidate reward function, solve RMAB problem using Whittle index method \citep{whittle1988restless} and generate trajectory outputs (see \cite{math11071639} or the Appendix in \cite{kim2025dpopro} for the explanation on the Whittle index), iii) sample 50 pairs of reward functions from this set and query the LLM-judge to select preferred reward function, iv) perform this query 10 times to estimate the LLM's uncertainty over preferences. Finally, we obtain a dataset of 9500 preferred and rejected reward function responses over 190 prioritization prompts.

\paragraph{Baselines}
In addition to the DPO variants, we also compare against the Decision Language Model (DLM), a self-reflection–based inference-time approach introduced in \cite{behari2024a} using Gemini 2.0. Note that DLM is not fine-tuned on the training data, so this method has no notion of noise. For each prompt, we simply apply the out-of-the-box Gemini model and run the self-reflection procedure described in \cite{behari2024a}. We use Gemini 2.0, as in the original work, rather than the open-source Phi-3B model used for fine-tuning. This is to assess how our method compares against a state-of-the-art closed model that a typical user would rely on, further strengthened by self-reflection.

\paragraph{Evaluation}
To evaluate the quality of each generated reward function, we follow a procedure parallel to the dataset construction. For every prompt in the evaluation set, we generate a reward function using a given fine-tuning method and obtain the corresponding RMAB policy using the Whittle Index. We repeat the same procedure using the chosen reward function associated with that prompt in the evaluation data, producing a reference policy for comparison.

We compare these policies using two complementary methods. First, because this task provides access to ground-truth human-designed reward functions from \citet{behari2024a}, we can directly evaluate RMAB policy quality by computing the achieved reward under this ground-truth function. We compare the average ground-truth reward obtained by the policy induced by the fine-tuned LLM with that obtained by the policy from the evaluation data. Second, we also perform an LLM-as-a-judge evaluation using GPT-4o-mini to compute win-rate as in Section \ref{sec:benchmarks-u0}. The judge compares the resulting trajectories of the RMAB policies, i.e., the resource allocation outcomes across the demographic data, to determine which policy is preferred.

\paragraph{Results}  

We report the results in Table~\ref{tab:combined-armman}. For the ground-truth reward metric, all variants of \textsc{DPO-PRO} outperform both DrDPO and Vanilla DPO in both the noiseless and high-noise settings. Their performance is slightly lower than DLM, but this gap must be interpreted in context: DLM requires (1) access to a high-end proprietary LLM API at deployment time and (2) substantially higher inference-time cost, as we show below.

For the win-rate metric, the results are more mixed. The \textsc{DPO-PRO} variants generally outperform DrDPO and Vanilla DPO in the low-noise setting, with the exception of the $\rho = 0.03$ variant. In particular, DPO-PRO ($\rho = 0.1$) outperforms both DrDPO and Vanilla DPO in both the noiseless and low-noise settings, while DPO-PRO ($\rho = 0.3$) performs similarly to these baselines under no noise and clearly exceeds them under low noise. As with the ground-truth reward metric, all \textsc{DPO-PRO} variants perform below DLM.

One interesting observation is that, unlike the clean trends in Section~\ref{sec:benchmarks-u0}, the performance across different $\rho$ values and noise levels in the ARMMAN task does not yield a simple or monotonic pattern. For example, larger $\rho$ values are not consistently more conservative, and performance even improves as noise is injected. Although pinpointing the exact cause is difficult, several factors likely contribute to this behavior.

First, unlike the curated benchmark data in Section~\ref{sec:benchmarks-u0}, the ARMMAN dataset is substantially noisier. Preference labels generated by LLM judges are difficult to obtain reliably for this task (see Section~\ref{sec:prelim}), reducing the quality of the signal available during fine-tuning. In other words, the noiseless setting here does not necessarily correspond to a clean preference signal, since the underlying annotations are already highly noisy. Second, DPO and related off-policy preference-based methods are known to be sensitive to overfitting and reward hacking when trained on fixed datasets \citep{pmlr-v238-gheshlaghi-azar24a,amini-etal-2024-direct,yang2024orthogonalfinetuningdirectpreference,xiao2025comprehensivesurveydirectpreference}. In such settings, injected noise can act as a regularizer: it discourages the model from fitting spurious or inconsistent preference patterns and promotes more stable decision rules. Similar trends, where performance improves under moderate noise, have been reported in \citep{wu2025drdpo,bukharin2024robust}. This interpretation is also consistent with recent theoretical work showing that controlled noise injection can improve generalization by implicitly encouraging simpler and more robust decision boundaries \citep{li2025towards}.


\paragraph{Inference Time Comparison}

\begin{figure}[tbp]
    \centering
    \includegraphics[width=0.8\linewidth]{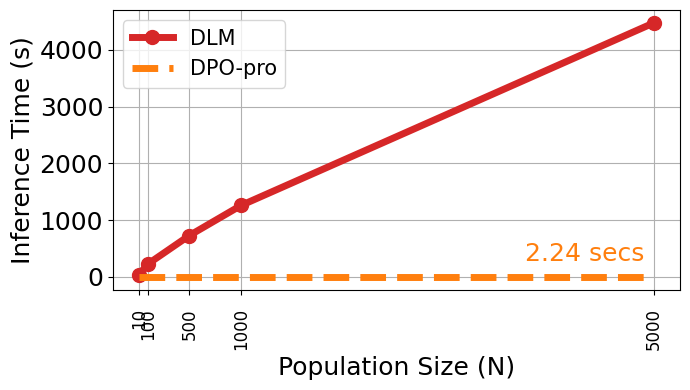}
    \caption{Inference-time comparison of \textsc{DPO-PRO} and DLM across different population sizes (i.e., number of arms) in the underlying RMAB.}
    \label{fig:inf}
\end{figure}

Figure~\ref{fig:inf} compares the inference time of \textsc{DPO-PRO} (with $\rho = 0.1$) and DLM. Unlike DLM, whose inference time scales linearly with the number of arms due to repeated simulation of resource-allocation policy and self-reflection, \textsc{DPO-PRO} incurs a constant inference cost regardless of problem size. This efficiency gap becomes increasingly significant in large-scale settings, making \textsc{DPO-PRO} more scalable. While overall performance is comparable or mixed across evaluation metrics, minimal inference-time overhead makes \textsc{DPO-PRO} especially well-suited for real-world deployment.

\paragraph{Examples}

To further illustrate the differences between vanilla DPO and \textsc{DPO-PRO} (with $\rho = 0.1)$, we present response outputs from both models on two representative tasks from the evaluation dataset (Table~\ref{tab:armman_reward_funs}). Many tasks contain attributes that can be interpreted in multiple ways. For instance, the concept of ``young beneficiaries'' can be identified directly through the age feature, but also indirectly via lower education level or income. Similarly, a task that expresses a preference for ``midday calls'' could refer explicitly to the 12:30–3:00 PM time slot, but might also include the 10:30–12:30 PM window, depending on interpretation.

In such cases, we observe that vanilla-DPO favors uncommon interpretations of ambiguous attributes based on spurious correlations in noisy preference data. In contrast, \textsc{DPO-PRO} adopts a cautious strategy by selecting narrower definition, aiming to be robust to uncertainty in the preference signal at evaluation.

\begin{table}[t]
\small
\centering

\begin{subtable}[t]{.99\columnwidth}
\centering
\caption{Average Ground Truth Reward ($\uparrow$).}
\scalebox{0.85}{
\begin{tabular}{lccc}
\toprule
Method & No noise & Low noise \\
\midrule
DLM                     & $6073.04$ & \textit{NA} \\
DPO            & $5465.36$ & $5707.09$ \\
DrDPO                  & $5480.06$ & $5675.17$ \\
DPO-PRO ($\rho=0.008$)  & $5530.58$  &  $5832.66$ \\
DPO-PRO ($\rho=0.03$)  & $5557.9$  & $5790.8$  \\
DPO-PRO ($\rho=0.1$)   & $5677.76$ & $5807.83$ \\
DPO-PRO ($\rho=0.3$)   & $5797.1$  & $5904.7$  \\
\bottomrule
\end{tabular}}
\end{subtable}


\begin{subtable}[t]{.99\columnwidth}
\centering
\caption{Win rate (\%, $\uparrow$).}
\scalebox{0.85}{
\begin{tabular}{lccc}
\toprule
Method & No noise & Low noise \\
\midrule
DLM                     & 24.82 & \textit{NA} \\
DPO            & 15.38 & 13.84 \\
DrDPO                  & 15.89 & 12.13 \\
DPO-PRO ($\rho=0.008$)  & 14.18  & 14.87 \\
DPO-PRO ($\rho=0.03$)  & 13.5  & 13.5 \\
DPO-PRO ($\rho=0.1$)   & 17.7  & 14.35 \\
DPO-PRO ($\rho=0.3$)   & 15.7  & 17.8 \\
\bottomrule
\end{tabular}}
\end{subtable}

\caption{Results on ARMMAN environment under no-noise and low-noise settings.}
\label{tab:combined-armman}

\end{table}

\begin{table}[t]
\renewcommand{\arraystretch}{1.05}
\centering
\scalebox{0.82}{
\begin{tabular}{@{}l@{\hspace{0.5em}}p{8cm}@{}}
\toprule
\multicolumn{2}{@{}l}{\textbf{Task 1:} Prefer both young and elderly beneficiaries} \\ 
\midrule
Vanilla-DPO & 
\footnotesize\parbox[t]{8cm}{%
\texttt{s + 3 * (youngest\_age or second\_youngest\_age or oldest\_age) +}\\
\texttt{~~~~2 * (lowest\_education or second\_lowest\_education or third\_lowest\_education) +}\\
\texttt{~~~~(lowest\_income or second\_lowest\_income or third\_lowest\_income)}
} \\[1ex]
DPO-PRO &
\footnotesize\parbox[t]{8cm}{%
\texttt{s + 3 * (youngest\_age or second\_youngest\_age or oldest\_age) +}\\
\texttt{~~~~2 * (lowest\_education or second\_lowest\_education or third\_lowest\_education)}
} \\
\addlinespace[0.5ex] \midrule
\multicolumn{2}{@{}l}{\textbf{Task 2:} Prioritize midday calls and NGO-registered users} \\ 
\midrule
Vanilla-DPO & 
\footnotesize\parbox[t]{8cm}{%
\texttt{s + 3 * (10\_30-12\_30pm and NGO\_registered) +}
\texttt{~2 * (12\_30-3pm and NGO\_registered)}
} \\[1ex]
DPO-PRO &
\footnotesize\parbox[t]{8cm}{%
\texttt{s + 3 * (12\_30-3pm and NGO\_registered)}
} \\
\bottomrule
\end{tabular}
}
\caption{Examples of reward functions produced by Vanilla-DPO and DPO-PRO when trained under noise. DPO-PRO produces more conservative preference interpretations.}
\label{tab:armman_reward_funs}
\end{table}

\section{Conclusion}
\label{sec:conclusion}

Motivated by reward-function design in RMABs for public health, we introduced \textsc{DPO-PRO}, a robust fine-tuning algorithm that extends DPO by explicitly modeling uncertainty in preference distributions through a lightweight DRO formulation. Our method is particularly well-suited for high-stakes decision-making settings such as public health, where preference annotations are inherently noisy and often limited. Compared to existing DRO-based approaches, \textsc{DPO-PRO} avoids excessive conservatism while offering a theoretical interpretation as a regularized variant of DPO. Through extensive evaluation on real-world public health data and standard alignment benchmarks, we showed that \textsc{DPO-PRO} improves robustness and achieves performance competitive with inference-time methods at a fraction of their computational overhead. Moreover, compared with the previous inference-time approach, \textsc{DPO-PRO} eliminates the need for external LLM APIs at deployment, an essential property when handling sensitive or private data. These results highlight the potential of \textsc{DPO-PRO} as a scalable and reliable tool for learning aligned reward functions in real-world policy learning tasks for public health.

\section*{Acknowledgements}

This work was supported by ONR MURI N00014-
24-1-2742 and based upon work supported by the AI Research Institutes Program funded by the National Science Foundation under the AI Institute for Societal Decision Making (NSF AI-SDM), Award No. 2229881.

\bibliography{references}

\clearpage

\end{document}